\pdfoutput=1

\documentclass[11pt]{article}

\usepackage[preprint]{acl}
\usepackage{nert}

\usepackage{times}
\usepackage{latexsym}
\usepackage{kotex}
\usepackage{linguex}
\usepackage{hyperref}
\usepackage{booktabs}
\usepackage{xcolor}
\usepackage{expex}
\usepackage{CJKutf8}

\usepackage[T1]{fontenc}

\usepackage[utf8]{inputenc}

\usepackage{microtype}

\usepackage{inconsolata}

\usepackage{graphicx}

\title{A Curious Class of Adpositional Multiword Expressions in Korean}

\author{Junghyun Min\textsuperscript{1} \quad\quad
Na-Rae Han\textsuperscript{2} \quad\quad
Jena D. Hwang\textsuperscript{3} \quad\quad 
Nathan Schneider\textsuperscript{1}\\
\textsuperscript{1}Georgetown University \quad
\textsuperscript{2}University of Pittsburgh \quad
\textsuperscript{3}Allen Institute for AI\\
  \{\emldisplay{jm3743@georgetown.edu}{jm3743}, \emldisplay{nathan.schneider@georgetown.edu}{nathan.schneider}\}\texttt{@georgetown.edu}\\ \eml{naraehan@pitt.edu} \quad \eml{jenah@allenai.org}
}

\newcommand{\suff}[1]{\mbox{\textit{#1}}} 

\begin{document}
\maketitle
\begin{abstract}
Multiword expressions (MWEs) have been widely studied in cross-lingual annotation frameworks such as PARSEME. 
However, Korean MWEs remain underrepresented in these efforts.
In particular, Korean multiword adpositions lack systematic analysis, annotated resources, and integration into existing multilingual frameworks.
In this paper, we study a class of Korean functional multiword expressions: postpositional verb-based constructions (PVCs).
Using data from Korean Wikipedia, we survey and analyze several PVC expressions and contrast them with non-MWEs and light verb constructions (LVCs) with similar structure.
Building on this analysis, we propose annotation guidelines designed to support future work in Korean multiword adpositions and facilitate alignment with cross-lingual frameworks. 
\end{abstract}

\section{Introduction}
\label{sec:intro}

PARSing and Multiword Expressions \citep[PARSEME; ][]{savary2015parseme} is a robust multilingual framework for annotating idiomatic word combinations, known as multiword expressions (MWEs).
They are distinguished from literal and fully productive combinations via a suite of linguistic tests, and are classified into subcategories based on their grammatical structure.
Similarly to words, MWEs are categorized according to their morphosyntactic and functional criteria \citep{baldwin2010multiword, savary-etal-2023-parseme}.

One category of MWEs is expressions that act like prepositions or postpositions (collectively `adpositions').
The PARSEME~2.0 guidelines encompass a wide range of MWE types, including adpositional MWEs as a subtype of functional MWEs\footnote{\url{https://parsemefr.lis-lab.fr/parseme-st-guidelines/2.0/}}.
An English example (sometimes termed a \textit{complex preposition}) is \p{in front of}.
With respect to its syntactic distribution, \p{in front of} is similar to single-word adpositions like \p{behind} and \p{near}. 
It also carries a relational, and in particular spatial, meaning typical of adpositions.
Finally, it is an MWE by PARSEME guidelines since it exhibits grammatical fixedness (fossilization): it has been lexicalized to not allow morphological changes (*\w{in fronts of}) or modifiers (*\w{in far front of}).

In this paper, we consider adpositional MWEs in Korean against the backdrop of the PARSEME framework.
We focus on a pattern of grammaticalized adpositional MWEs that we term \textbf{postpositional verb-based constructions (PVCs)}. An example appears in (\ref{kwanhan-attr}):

\ex
\small
\begingl
\gla 게\p{에} \p{관한} 책 // 
\glb key\p{-ey} \p{kwanha-n} chayk //
\glc crab-OBL relate-ADN book  //
\glft `a book about crabs' //
\endgl
\label{kwanhan-attr}
\xe
In (\ref{kwanhan-attr}), the semantically bleached postposition \p{\mbox{-에~ey}}  combines with the verb \vpred{관한 kwanhan}---fossilized in its adnominal (or attributive\footnote{`Attributive' contrasts with predicative; `adnominal' focuses on its function as a modifier of a noun. Both refer to the same set verb endings in Korean.}) form---to constitute the adpositional MWE \p{-에 관한 ey kwanhan} `about'.\footnote{Throughout this paper, we format \p{adpositions} (including PVCs), \vpred{verbs}, and \suff{other categories, including suffixes}.} As a unit, the MWE marks the topic argument of the head nominal `book' to mean `book \p{about} crabs.'

To the best of our knowledge, this is the first study of PVCs.
Prior studies of Korean have examined a range of classes of MWEs and idiomatic expressions, including a range of adpositional expressions, and the polysemy of grammatical markers that come into play in PVCs (\cref{sec:background}).
Moreover, adpositions and adpositional MWEs have been semantically annotated in the multilingual SNACS framework \citep[\emph{inter alia}]{schneider-etal-2018-comprehensive, schneider2022adpositioncasesupersensesv26,arora-etal-2021-snacs}, but the Korean implementation of SNACS \citep{hwang-etal-2020-k} has not annotated adpositional MWEs.\footnote{And likewise for Japanese, which is structurally similar \citep{aoyama-etal-2024-j}.}

In this paper, we develop an account of this construction and a proposal for how its instances should be annotated in the PARSEME framework. We contribute:
\begin{itemize}
\item an initial list of such Korean PVCs, drawn from Korean Wikipedia;
\item a proposal for how to annotate them under current PARSEME guidelines; and
\item a proposed linguistic analysis of PVCs, including a discussion on distinguishing them from adnominal verbs.
\end{itemize}

\begin{table*}[ht]
\centering
\small
\begin{tabular}{lHlHllH}
\toprule
\textbf{Adposition} & Total Occ. & \textbf{Bound stem} & Main pred.? & \textbf{Meaning} & \textbf{Suffix forms with \p{하다 hata} inflections} & Homonyms \\
\midrule
-에 \textit{ey}; oblique & 272,022 & 대 \textit{tay} & FALSE & about & -한 \textit{han}, -해 \textit{hay}, -해서 \textit{hayse}, -하여 \textit{haye} & 를 대하다 \textit{ul tayhata} ‘treat’ \\

-에 \textit{ey} & 170,110 & 의 \textit{ui} & FALSE & by & -한 \textit{han}, -해 \textit{hay}, -해서 \textit{hayse}, -하여 \textit{haye} & \\

\suff{-를 lul}; accusative & 122,731 & 통 \textit{thong} & FALSE & via, through & -한 \textit{han}, -해 \textit{hay}, -해서 \textit{hayse}, -하여 \textit{haye} & 와 통하다 \textit{wa thonghata} ‘connect, flow’ \\

\suff{-를 lul} & 119,673 & 위 \textit{wi} & FALSE & for & -한 \textit{han}, -해 \textit{hay}, -해서 \textit{hayse}, -하여 \textit{haye} & 를 위하다 \textit{lul wuihata} ‘care for’ \\

-로 \textit{lo; dative} & 85,339 & 인 \textit{in} & FALSE & due to & -한 \textit{han}, -해 \textit{hay}, -해서 \textit{hayse}, -하여 \textit{haye} & \\

-에 \textit{ey} & 63,819 & 관 \textit{kwan} & FALSE & about & -한 \textit{han}, -해 \textit{hay}, -해서 \textit{hayse}, -하여 \textit{haye} & \\

-에 \textit{ey} & 38,378 & 속 \textit{sok}\textsuperscript{\dagger} & TRUE* & in, belong to & -한 \textit{han}, -해 \textit{hay} & \\

-로 \textit{lo}, 를 \textit{lul} & 19,777 & 향 \textit{hyang}\textsuperscript{\dagger} & TRUE* & towards & -한 \textit{han}, -해 \textit{hay} & \\

-에 \textit{ey} & 18,449 & 비 \textit{pi} & FALSE & than, compared to & -한 \textit{han}, -해 \textit{hay}, -해서 \textit{hayse}, -하여 \textit{haye} & 와 비하다 \textit{wa pihata} ‘be comparable to’ \\

-에도 \textit{eyto}; oblique + additive & 14,989 & 불구 \textit{pwulkwu} & FALSE & although & -하고 \textit{-hako} & 불구되다, 불구가 되다 \\

\suff{-를 lul} & 12,233 & 비롯 \textit{piros} & FALSE & such as & -한 \textit{han}, -해 \textit{hay}, -해서 \textit{hayse}, -하여 \textit{haye} & 에서 비롯하다, 비롯되다 \\

\suff{-를 lul} & 4,839 & 기 \textit{ki}\textsuperscript{\dagger} & TRUE* & since & -한 \textit{han}, -해 \textit{hay} & \\

-에 \textit{ey} & 3,945 & 반 \textit{pan} & FALSE & against, unlike & -한 \textit{han}, -해 \textit{hay}, -해서 \textit{hayse}, -하여 \textit{haye} & 에(게) 반하다 ‘fall for’ \\

\suff{-를 lul} & 135 & 위시 \textit{wisi} & FALSE & such as & -한 \textit{han}, -해 \textit{hay}, -해서 \textit{hayse}, -하여 \textit{haye} & \\
\bottomrule
\end{tabular}
\caption{
Non-exhaustive list of distributional properties of 14 Korean PVCs, in order of frequency in Korean Wikipedia. Additional adpositions and suffix forms are possible for select arguments as described in \cref{sec:korean-PVCs}. \suff{-해 -hay}, \suff{-해서 hayse}, \suff{-하여 haye} are resultative connective suffixes; \suff{-하고 hako} is an conjunctive connective suffix; \suff{-한 han} is an adnominal suffix. \textsuperscript{\dagger}Unlike others, these bound stems can serve as the main predicate in a sentence, as discussed in \cref{sec:attr-conn}.}

\label{tab:korean-pvcs}
\end{table*}

\section{Defining PVCs}
Single-word Korean adpositions are postpositions called 조사 \textit{cosa} \footnote{Pronounced `josa'.} that are orthographically and phonologically bound to a noun phrase \citep{martin1992reference, sohn2001korean, yeon2019korean}.
They function as case markers, conjunctions, and markers of discourse-pragmatic meaning like topic and focus \citep{hwang-etal-2020-k}.

We define Korean PVCs as adpositional multiword units composed of two parts: a \textit{postposition} that attaches to a noun phrase and a verb that undergoes limited inflection.\footnote{While we do not discuss Korean adpositional MWEs beyond PVCs, we are aware of several postpositional units that may be considered multiword expressions, like stacked postpositions \citep{hwang-etal-2020-k} and other postposition equivalents \citep{moon2015constructions}.}

This paper focuses on \suff{하다 -hata} PVCs, whose verb portion is composed of a \textit{bound stem} and a \textit{verbalization suffix} \suff{하다 -hata} that attaches to the bound stem.\footnote{PVCs also extend to constructions that do not contain the \suff{-hata} suffix such as \p{-에 따르면 -ey ttalumyen} `according to' containing a bound stem lexicalized from the verb \vpred{따르다 ttaluta} `to follow'. As with \suff{하다 -hata} PVCs, the verb is limited in inflection.}
We thus describe them as having three components: a postposition, a bound stem, and \suff{하다 -hata} suffix.
Together, they mark the relationship between the head and the object noun. As illustrated in (\ref{kwanhan-attr}), the adposition \p{-에 ey} and the bound stem \vpred{대 tay} with the verbalization suffix in the adnominal form (\suff{-한 han}) comprise a PVC that corresponds to the English preposition \p{about}.

A key characteristic of Korean postpositional MWEs is that they are fossilized in meaning. 
Although the adnominal (e.g., \vpred{관한 kwanhan}) in example (\ref{kwanhan-attr}) is in the attributive form of a verb (e.g., \vpred{관하다 kwanhata} `to relate to'), it cannot be productively used as a matrix verb of a sentence and does not participate in regular verbal inflection as exemplified in (\ref{kwanhan4-matrix}).
Additionally, it cannot freely undergo regular morphological change as shown in (\ref{kwanhan4-morph}), appearing almost exclusively within a limited set of constructions allowed for the MWE, further discussed in \cref{sec:attr-conn}.
%
\ex
\small\judge*
\begingl
\gla 책이 게\p{에} \p{관했다} // 
\glb chayk-i key\p{-ey} \p{kwanha-yss-ta}  //
\glc book-NOM crab-OBL relate-PAST-DECL  //
\glft `A book was about crabs.' //
\endgl
\label{kwanhan4-matrix}
\xe
\vspace{-2em}
\ex
\small\judge
\begingl
\gla 게\p{에} \p{관한~/~*했던} 토론 // 
\glb key\p{-ey} \p{kwanha-n~/~*-yss-ten} tholon //
\glc crab-OBL relate-ADN~/~-PAST-ADN debate  //
\glft `a debate that used to be about crabs' //
\endgl
\label{kwanhan4-morph}
\xe

This is in contrast to the full predicate verbs with suffix \vpred{-hata}. Example (\ref{kuhan-attr-1}) shows the predicate \vpred{구하다 kuhata} `to rescue' appearing in the same adnominal construction as PVCs. But the key distinguishing factor is that this can also be freely used as a matrix verb with a full range of inflectional endings 
while retaining its core meaning as seen in example (\ref{kuhan-attr-1}).
\ex
\small
\begingl
\gla 친구{를} \p{{구한}~/~했던}  강아지 // 
\glb chinku\p{-lul} \p{kuha-n~/~-yss-ten} kangaci //
\glc friend-ACC rescue-ADN~/~-PAST-ADN puppy  //
\glft `a puppy that rescued a friend' //

\endgl
\label{kuhan-attr-1}
\xe

\noindent These characteristics notably pass two of the tests for fixedness under the PARSEME guidelines \citep{savary2015parseme}.



\section{Extracting a list of PVCs}
\label{sec:finding-pvcs}



A range of expressions fall in the PVC class. To compile an initial list of these, we develop a pipeline to extract and filter candidates from corpora.
The main challenge is disambiguating them from false positives (verbs and light verb constructions appearing in the same adnominal construction as PVCs, as in (\ref{kuhan-attr-1})).

\paragraph{Candidate extraction} 
First, we take the May 2024 dump of Korean Wikipedia totaling 515k articles \citep{korean_wikipedia, wikidump}.
Then, we analyze the main text of each article with \texttt{konlpy}'s \citep{park2014konlpy} \texttt{Mecab} morphological analyzer \citep{kudo2013mecab}.
Finally, using a regular expression, we look for sequences of an adposition (with a \texttt{J*} tag), a bound stem (\texttt{XR} but commonly erroneously parsed as a noun \texttt{NN*}), and some inflection of suffix \suff{-hata}.\footnote{We release our pipeline at \url{https://github.com/aatlantise/korean-multiword-postpositions}.}

For each stem, we retrieve a list of adpositions, suffixes, and adposition-stem-suffix sequences that occur with it.
As PVCs are more lexicalized than the usual verb-argument construction, bound stems that form PVCs co-occur with smaller numbers of adpositions, suffixes, and sequences, which may be used to select bound stems that can form PVCs.
However, to ensure accurate retrieval, we take the 300 most frequently occurring stems in such sequences out of 3.5k candidates and manually verify whether the stem forms a PVC.

\paragraph{Annotating candidates}

To verify the PVC-hood of a candidate, we exploit the properties of PVCs described in \cref{sec:korean-PVCs} and connected to those of functional MWEs \citep{savary2015parseme}.
PVC verbs:
\begin{itemize}
    \item exhibit limited capacity for morphological inflection (see (\ref{kwanhan4-morph})),
    \item cannot be modified (\ref{kwanhan3-mod}), and
    \item do not serve as a main predicate of a sentence or a clause (discussed further in \cref{sec:attr-conn}).
    Exceptions exist, which we do not treat as instances of PVCs; see \cref{sec:predicative-pvc-verbs}.
\end{itemize}

Among the 300 stems, 12 are bound stems that form PVCs, and the rest are verbs (mostly light verb constructions) with cased arguments.
Of the 12, 3 stems can also serve as a main predicate of a sentence or a clause; they only form PVCs when they do not (\cref{sec:attr-conn}).
Our additional manual analysis uncovers 2 additional, less frequent bound stems that form PVCs.

\section{Properties of PVCs}
\label{sec:korean-PVCs}

The 14 identified PVCs, their components, and their meanings appear in \cref{tab:korean-pvcs}, sorted from most to least frequent in Korean Wikipedia.
The list of applicable postpositions and suffixes is not exhaustive. For example, 
\p{-로 향한 lo hyanghan} `towards'
can surface with 
\p{-에게 eykey} when it attaches to animate arguments (\p{-에게 향한 eykey hyanghan}).

We verify that PVCs are indeed MWEs and adpositions, in accordance with the PARSEME guidelines \citep{savary2015parseme}.
Distributionally, they can be replaced with single-word adpositions: for example, \p{-를 위한 lul wihan} `for' can be replaced by \p{-의 ui} `of' or \p{-같은 kath'un} `like' while retaining grammaticality and acceptability.

As expected, Korean PVCs exhibit morphological or lexical inflexibility that characterize MWEs. 
Examples (\ref{kwanhan4-matrix}), (\ref{kwanhan4-morph}), and (\ref{kwanhan3-mod}) show that modification of a component (the verb) and regular morphological change into the past tense results in questionable acceptability, where both examples are better analyzed as a predicate in a relative clause.
We note that despite the lexicalization, PVCs exhibit neither irregular syntactic structure nor semantic idiomaticity; we thus describe them as weak MWEs \citep{schneider2014discriminative}.
\ex
\small\judge*
\begingl
\gla 게\p{에} 약간 \p{관한} 책 // 
\glb key\p{-ey} yakkan \p{kwanha-n} chayk //
\glc crab-OBL somewhat relate-ADN book  //
\glft `a book somewhat about crabs' //
\endgl
\label{kwanhan3-mod}
\xe

\subsection{Adnominal and connective forms only}
\label{sec:attr-conn}

A distinctive property of Korean PVCs, resulting from their lexicalization, is that the verb suffix \suff{-하다 hata} present in the PVC can only be inflected in limited cases discussed below. 

Unlike regular verbs, many that form PVCs entirely resist appearing as a tensed matrix predicate: progressive \vpred{*관한다 kwanhanta} and past \vpred{*관했다 kwanhayssta} would be ungrammatical for (\ref{kwanhan-attr}). For others, doing so forces a change in meaning: the PVC \p{-에 대한 ey tayhan} means `about', but \vpred{-을 대했다 ul tayhayssta} in the past-tense matrix form require a different postposition \textit{and} changes the meaning to `treated'. 

As a result, \suff{-hata} suffixes in PVCs have a constrained distribution tied to specific inflectional endings: they only appear in a pre-nominal position with the adnominal ending \suff{-한 han} (shown in (\ref{kwanhan-attr})) or in a pre-verbal position with connective endings (\suff{-해 hay}, \suff{-해서 hayse}, \suff{-하여 haye})\footnote{The connective suffix forms are generally analyzed in the literature to form serial verb constructions that together denote a single event \citep{kim2010serial, im-lee-2001-type}.}.

A small set of PVCs exhibit exceptions: \vpred{속하다 sokhata} `be a part of', \vpred{향하다 hyanghata} `face, head towards', and \vpred{기하다 kihata} `set as time for something to begin'. These permit predicative use without a major semantic shift. However, we treat these as PVCs, as the predicative forms are generally dispreferred in favor of their respective PVC forms or alternate constructions. See \cref{sec:predicative-pvc-verbs} for more discussion.

\subsection{PVC verbs are not light verb constructions}
\label{sec:not-lvcs}
In addition to the predicate verb and verbal suffix, \vpred{하다 hata} `to do' can function as a light verb that combines with a noun complement to become lexicalized and form light verb constructions \citep[example (\ref{sanchayk-lv}); ][]{chae-1996-light, han2000predicates}. In fact, just like the full predicate, these LVCs can also participate in verbal inflections. As a result, verbs in PVCs like \p{관한 kwanhan} (example (\ref{kwanhan-attr})) can resemble LVCs, especially when the LVC appears as an adnominal. 
\ex
\small
\begingl
\gla 공원을 산책하다//
\glb kongwuen-lul sanchayk-ha-ta //
\glc park-ACC stroll-LV-DECL//
\glft `take a stroll at the park' //
\endgl
\label{sanchayk-lv}
\xe
\ex
\small
\begingl
\gla 공원을 산책한 사람//
\glb kongwuen-lul sanchayk-ha-n salam //
\glc park-ACC stroll-LV-ADN person//
\glft `person that took a stroll at the park' //
\endgl
\label{sanchayk-lv-adn}
\xe
In an LVC, most of the semantic content  is contributed by the noun \citep{jespersen1965modern, cattell1984composite, baldwin2010multiword}.  The noun in an LVC like \w{산책 sanchayk} `stroll' is a fully unbound lexical item \citep{chae-1996-light}, such that can be reduced to a nominal form, as in (\ref{sanchayk2-nom}).
%
\ex
\small
\begingl
\gla 공원 산책//
\glb kongwuen sanchayk//
\glc park stroll//
\glft `a stroll at the park'//
\endgl
\label{sanchayk2-nom}
\xe
\ex
\small\judge*
\begingl
\gla 게 관//
\glb key kwan//
\glc crab relate//
\endgl
\label{relate-nom}
\xe
In a PVC, while it could be argued that the semantic content of \vpred{관하다 kwanhata} `relate' stems from \w{관 kwan},\footnote{\w{관 kwan} is a Sino-Korean stem from Middle Chinese \w{關 kwaen} meaning `relation' or `barrier' \citep{baxter2010handbook}.} PVC verb stems are bound and cannot appear on their own---as shown in example (\ref{relate-nom}), which is structurally analogous to (\ref{sanchayk2-nom}). 
They are comparable to English bound morphemes like \w{fer} in \vpred{refer} or \vpred{confer}. Thus, they are unable to be reduced to nominal forms, as required by PARSEME guidelines \citep{savary2015parseme}.


\subsection{Other constructions with same structure}
\label{sec:other-constructions}

In this section, we discuss how PVCs with predicative and non-predicative verbs ((\ref{kwanhan4-matrix}), (\ref{kwanhan4-morph}), and (\ref{kwanhan3-mod})) compare to LVCs (\ref{sanchayk-lv})
and non-MWEs (\ref{kuhan-attr-1}) with cased arguments, as they can form the same surface structure consisting of a noun with adposition, a stem (or noun), and a \suff{-hata} verbal suffix.
We consider 3 constructions: PVCs with non-predicative verbs (\textbf{PVC-n)}, PVCs with possibly predicative verbs (\textbf{PVC-p}; denoted with \textsuperscript{\dagger} in \cref{tab:korean-pvcs}), and verbs with cased arguments (\textbf{verb + arg}), in the order from most to least lexicalized.
We compare them in 4 forms: adnominal (attributive) verb (\ref{kwanhan-attr}), main predicative verb (e.g.~\vpred{향하다 hyanghata} `face, head towards' in \cref{sec:attr-conn}), verb modification (\ref{kwanhan3-mod}), and serial verb modification.

\Cref{tab:comparison} offers grammaticality judgments.
Despite the same surface structure across the 4 forms, the grammaticality of these forms varies across the 3 constructions.
PVCs with non-predicative verbs, comprising the majority of PVCs, are the most lexicalized, only able to appear in adnominal and connective forms.
While predicative verbs that form PVCs are more flexible, they are only PVCs when they do not appear as a main predicative verb, as they can be modified and undergo inflectional morphology as main predicates.
They are PVCs when they appear as the nonfinal component of a serialized verb.
We provide explicit examples and further explanations on grammaticality judgments of these constructions in \cref{appendix:examples}.

\begin{table}[t!]
\small
    \centering
    \begin{tabular}{rcccc}
    \toprule
        \textbf{Forms} & \textbf{PVC-n} & \textbf{PVC-p} & \textbf{verb + arg}\\
        \midrule
        Adnominal verb& $\checkmark$ & $\checkmark$ & $\checkmark$   \\
        Main predicate&  & $\checkmark$\textsuperscript{\dagger} & $\checkmark$   \\
        Adn.~verb mod.&  & $\checkmark$\textsuperscript{\dagger} & $\checkmark$   \\
        Conn.~verb mod. &  & & $\checkmark$   \\
        \bottomrule
    \end{tabular}
    \caption{Grammaticality of 3 categories of constructions with a PVC-like surface form. PVCs with non-predicative verbs are the most lexicalized, only able to appear in adnominal (attributive) forms.
    \textsuperscript{\dagger}Predicative verbs that form PVCs may undergo modification or serve as the main predicate in a sentence, but with such change they no longer carry adpositional meaning and instead behave like a regular verb-argument construction.}
    \label{tab:comparison}
\end{table}

\section{Related Work}
\label{sec:background}

We review studies of Korean MWEs and idiomaticity.

\paragraph{Traditional accounts.}
Traditional Korean grammar identifies several categories of idiomatic expressions. Examples include 
관용어 \textit{kwanyong-e} ``habitual word''; 
숙어    \textit{swuk-e}       ``familiar word''; and
연어    \textit{yen-e}       ``connected words, collocations''.
The boundaries of these categories tend to be fuzzy, with dictionaries disagreeing with each other.
See \cref{appendix:traditional-idioms} for further details.

\paragraph{Contemporary linguistic accounts.}
Linguistic accounts have offered taxonomical and theoretical analyses of Korean MWEs.
%
%
Prior work on Korean MWEs and LVCs describes their lexical and morphological inflexibility, analyzes LVC subtypes (e.g., common noun vs. serial verb constructions, often involving Sino-Korean nouns), and provides generative accounts of nominal–light verb combinations, noting the diverse realizations of light verbs and adjectives \citep{han-rambow-2000-sino, lee-2011-two, im-lee-2001-type, chae-1996-light, bak2012light}.
The form \vpred{하다 hata} `to do' has been a particular subject of study in its range of grammatical and semantic realizations, as a light verb, as a light adjective, or as a suffix \citep{han2000predicates}.
It is important to note that while verbs that appear in PVCs contain \suff{-하다 hata}, they are suffixes that attach to bound stems rather than light or support verbs that attach to nouns \citep{chae-1996-light, chae-2013-myths}.

\subparagraph{Postpositional MWEs.}
Early work on modern Korean discusses postpositional phrases and compound postpositions \citep{underwood1890introduction, roth1937han}.
More recent work following \citet{kim2002modifiers} has used the term \emph{postposition equivalents} for units that function like postpositions.
While many postposition equivalents are cased nouns denoting spatial relations \citep[e.g. \p{앞에 ap-ey} front-LOC `\p{at} the front, \p{in front of}'; ][]{suh2004constructions}, some Korean postposition equivalents have recently been described as adpositional multiword expressions that are metaphoric \citep{han2024multiword}, although the authors focus on (experiential) metaphoricity and idiomaticity of these MWEs rather than their MWE-hood, composition, or structure.
Other postposition equivalents have been analyzed as  collocations or ongoing processes of grammaticalization  \citep{kim2002modifiers, moon2015constructions}.
No prior work has focused on PVCs.

\paragraph{Korean MWE processing for machine translation.}
Outside of linguistics, work in machine translation has discussed Korean MWEs, including the PVCs we discuss in this paper, despite inconsistent terminology.
In their study of the challenges of English \textit{suk-e} in English-Korean machine translation, \citet{lee1993english} discuss English MWEs and attempt to create a dictionary mapping each of them to Korean counterparts.
Finally, in a study of Korean-to-Japanese multiword ``translation units,'' \citet{moon-lee-2000-representation} include a discussion on ``semi-words'' \p{-를 위한 lul wihan} `for' and \p{-のための no tameno} (`for'), the former of which we describe as a PVC in this paper.

\section{Conclusion}
In this paper, we describe a class of Korean multiword adpositions: postpositional verb-based constructions (PVCs).
We offer a list of 14 PVCs extracted from a corpus (\cref{tab:korean-pvcs}), analyze their verbs and their distributions (\cref{sec:attr-conn}, \cref{sec:not-lvcs}), and compare them to similar constructions with the same surface form (\cref{sec:other-constructions}).
This work is the first of its kind to analyze Korean multiword adpositions and draw a connection to cross-lingual annotation frameworks like PARSEME.

\section*{Limitations and future work}
We have highlighted postpositional verb-based constructions (PVCs) in Korean and examined their linguistic properties towards annotating them in the PARSEME framework.
Future work should pursue comprehensive annotation of Korean adpositional MWEs---including PVCs---in corpus data.
This will shed light on properties such as ambiguity and fossilization.

Our analysis is not exhaustive: we note distributional variation across domains. For example, the gerundial \suff{-함 ham} suffixes that denote 
a state are frequently attested in legal text. Our analysis will benefit from future work on such variation.

Our corpus analysis relies on the \texttt{Mecab} morphological analyzer 
\citep{kudo2013mecab}. While efficient and effective, the analyzer is not error-free; we thus do not report various metrics we measure in this paper, including the number of adpositions, suffixes, and sequences with which each stem occurs. The metrics are still available on our codebase.

In addition, comparative analysis with similar Japanese constructions formed with a postposition, Sino-Japanese stem, and an inflection of Japanese suffix \suff{-する suru} (e.g., 
\begin{CJK}{UTF8}{min}
\small
\p{-に対して}
\end{CJK}
\p{ni taishite} `about') may yield fruitful insights.
Note Sino-Japanese character 
\begin{CJK}{UTF8}{min}
\small
\p{対}
\end{CJK}
corresponds to \w{대 tay} in \p{-에 대한 ey tayhan} `about' from Sino-Korean 對.

\section*{Ethics statement}
In our research, we made use of publicly available datasets published on the web.
We acknowledge that the data obtained from the web may contain potential biases. 
We ensured that all datasets employed in our study were accessed and used in a manner that respects their intended use and complies with any associated licenses or terms of service.

We disclose our use of Gemini\footnote{\url{https://gemini.google.com/}} as a coding assistant.
We acknowledge the environmental and ethical considerations associated with the use of such AI technology, and have thoroughly reviewed the content to ensure that it does not include any unethical material.

\section*{Acknowledgements}
This paper is based on a project for the Georgetown University course ``All About Prepositions,'' taught by Nathan Schneider.
We thank Youngjin Kim and Youngbin Noh for their discussions on native speaker acceptability judgments; Wesley Scivetti, Yujin Seo, and students in All About Prepositions for their helpful comments on the project; and Amir Zeldes and Kohei Kajikawa for their help with Japanese analogs.
This research was supported in part by NSF award IIS-2144881.

\bibliography{custom}

\appendix

\section{Traditional approaches to Korean idiomatic expressions}
\label{appendix:traditional-idioms}
Traditional frameworks in Korean grammar comment on the use and history of idiomatic expressions.
Following \citeposs{sag2002multiword} definition of the multiword expression as ``lexical items that: (a) can be decomposed into multiple lexemes; and (b) display lexical syntactic, semantic, pragmatic, and/or statistical idiomaticity,'' we discuss the following terms: 
관용어 \textit{kwanyong-e} ``habitual word''; 
숙어    \textit{swuk-e}       ``familiar word'';
성어 \textit{seng-e}      ``word from the old times'';
연어    \textit{yen-e}       ``connected words,'' `collocations';
 속담   \textit{sokdam}      ``earthly talk,'' `proverbs'.

Korean \textit{yen-e} and \textit{sokdam} exhibit categorical differences, as they can respectively be compared to collocations and proverbs \citep{Lim2011Collocation, hoang2022vietnamese}, although \citet{Shim2009Korean} describes ``syntactic \textit{kwanyong-e}'' to be ``words that must go with each other'': collocations.

\citet{Doosan2003Dict} defines \textit{kwanyong-e} as ``expressions used habitually by the ordinary person'', \textit{swuk-e} as ``expressions that comprise of two or more words frequently used to function as a single word,'' and \textit{seng-e} as ``frequently quoted expressions since the old times.'' Despite distinct definitions offered, there is no clear categorical difference between the terms. Some dictionaries equate the terms \textit{kwanyong-e}, \textit{suk-e}, and \mbox{\textit{seng-e}} \citep{KNLI1999Dict, Doosan2003Dict}; others explicitly differentiate them. 
Doosan Encyclopedia provides English examples in its entries of the terms. Example of a \textit{kwanyong-e} is ``It rains cats and dogs'' and ``He kicked the bucket'' \citep{doosan2026kwanyonggu} while an example of \textit{suk-e} is ``get in'' \citep{doosan2026sugeo}, whose function as a single word is emphasized.
From this, we infer the encyclopedia categorically distinguishes the highly idiomatic ``cats and dogs'' and ``kick the bucket'' from other shorter, less idiomatic light verb constructions, prepositional verbs, and verb-particle constructions. 
On the other hand, a \textit{seng-e} is explained to usually have a Chinese origin, possibly due to influence from its subcategory \textit{saca-seng-e}, old 4-letter Chinese proverbs \citep{giles1873dictionary, Choi2016Vietnamese}.
Overall, we find that while there are several terms that can refer to multi-word expressions, their boundaries are often unclear and fuzzy to the point that dictionaries disagree with each other.

\section{Exceptions to PVC syntactic limitations}
\label{sec:predicative-pvc-verbs}

There are three notable exceptions to the limitations that verbs that form PVCs take: \vpred{속하다 sokhata} `be a part of', \vpred{향하다 hyanghata} `face, head towards', and \vpred{기하다 kihata} `set as time for something to begin' as outlined in \cref{tab:korean-pvcs}. 
These verbs can participate in the predicative from without deep change in meaning (\ref{hyang1-mod-morph}). 
We treat these as PVCs due to their restricted distribution and the general preference for corresponding PVC forms or alternative constructions (\ref{hyang2-serial}).
\ex
\small
\begingl
\gla 하늘로 갑자기 향했다 // 
\glb hanul-lo kapcaki hyangha-yss-ta //
\glc sky-DAT suddenly head-PAST-DECL  //
\glft `suddenly headed towards the sky' //
\endgl
\label{hyang1-mod-morph}
\xe
\ex
\small
\begingl
\gla 하늘\p{로} \p{향해} 날아가다 // 
\glb hanul-\p{lo} \p{hyangha-y} nalaka-ta //
\glc sky-DAT head-CONN fly-DECL  //
\glft `fly towards the sky' //
\endgl
\label{hyang2-serial}
\xe

\section{Grammaticality judgments of PVCs and related constructions in various forms}
\label{appendix:examples}
In this section, we provide glossed examples of various constructions and their grammaticality judgments discussed in \cref{tab:comparison} and \cref{sec:other-constructions}.
For each of the 3 constructions, we assign an event to be consistent in this section.
The non-predicative PVC example will continue featuring the book about crabs (\ref{kwanhan-attr} et al.).
The predicative PVC example will feature a ball headed towards the sky (\ref{hyang1-mod-morph}, \ref{hyang2-serial}).
The regular verb-argument construction examples will feature a puppy that rescued a friend.
All three verb lemmas are similar in appearance with a single-character bound stem and a \textit{-hata} suffix: \vpred{관하다 kwanhata} `relate,' a non-predicative PVC verb (\textbf{PVC-n)}, \vpred{향하다 hyanghata} `go towards,' a possibly predicative PVC verb (\textbf{PVC-p}), and \vpred{구하다 kuhata} `rescue,' a regular verb with a cased argument.

\paragraph{Adnominal forms.}
All three verbs are able to take the adnominal form (\ref{kwanhan-attr-app}--\ref{kuhan-attr}). 
\ex
\small
\begingl
\gla 게\p{에} \p{관한} 책 // 
\glb key\p{-ey} \p{kwanha-n} chayk //
\glc crab-OBL relate-ADN book  //
\glft `a book about crabs' [PVC-n] //
\endgl
\label{kwanhan-attr-app}
\xe
\ex
\small
\begingl
\gla 하늘\p{을} \p{향한} 공//
\glb hanul-\p{ul} \p{hyangha-n} kong //
\glc sky-ACC face-ADN ball//
\glft `a ball (headed) towards the sky' [PVC-p] //
\endgl
\label{hyanghan-attr-app}
\xe
\ex
\small
\begingl
\gla 친구\p{를} \p{구한} 강아지 // 
\glb chinku\p{-lul} kuha-n kangaci //
\glc friend-ACC rescue-ADN puppy  //
\glft `a puppy that rescued a friend' [verb+arg] //
\endgl
\label{kuhan-attr}
\xe

\paragraph{Main predicate forms.}
The non-predicative PVC's verb is unable to serve as a main predicate of a sentence (\ref{kwanhan-pred-app}). While the predicative PVC's can (\ref{hyanghata-pred}), it no longer forms a multiword expression, and behaves like the verb-argument construction.

\ex
\small\judge*
\begingl
\gla 책이 게\p{에} \p{관했다} // 
\glb chayk-i key\p{-ey} \p{kwanha-yss-ta}  //
\glc book-NOM crab-OBL relate-PAST-DECL  //
\glft `A book was about crabs.' [PVC-n] //
\endgl
\label{kwanhan-pred-app}
\xe
Example (\ref{kwanhan-pred-app}) is generally grammatically unacceptable.
However, in legal texts, the same construction is acceptable and even frequent when the verb takes the gerundial \textit{-함 ham} suffix (e.g.~to signal a state in which a book is about crabs).

\ex
\small
\begingl
\gla 공\p{이} 하늘\p{을} {향했다} //
\glb kong-\p{i} hanul-\p{ul} {hyangha-yss-ta}  //
\glc ball-NOM sky-ACC face-PAST-DECL //
\glft `A ball headed towards the sky.' [PVC-p] //
\endgl
\label{hyanghata-pred}
\xe
While grammatically acceptable, the verb in (\ref{hyanghata-pred}) carries predicative meaning instead of the adpositional meaning carried by (\ref{hyanghan-attr-app}).
In this usage, then, \vpred{향하다 hyanghata} `to head' is treated as a full lexical verb akin to the non-PVC verb \vpred{구하다 kuhata} `to rescue' (below in \ref{kuhata-pred}), which is subject to no restriction as a matrix verb.

\ex
\small
\begingl
\gla 강아지\p{가} 친구\p{를} {구했다} // 
\glb kangaci-\p{ka} chinku-\p{lul} kuha-yss-ta  //
\glc puppy-NOM friend-ACC rescue-PAST-DECL   //
\glft `A puppy rescued a friend.' [verb+arg] //
\endgl
\label{kuhata-pred}
\xe

\paragraph{Adnominal verb modification.}
Attempting to modify the verb shows that PVC verbs are not able to undergo modification without ungrammaticality (\ref{kwanhan-attr-mod1}) or shift in meaning (PVC to regular verb with argument; \ref{hyanghata-attr-mod1}). The PVC as a whole may be freely modified (\ref{kwanhan-attr-mod2}, \ref{hyanghata-attr-mod2}). The regular verb may be modified with or without the argument freely (\ref{kuhata-mod}).
\ex
\small\judge*
\begingl
\gla 게\p{에} 명백히 \p{관한} 책 // 
\glb key\p{-ey} myengpaykhi \p{kwanha-n} chayk //
\glc crab-OBL clearly relate-ADN book  //
\glft `a book clearly about crabs' [PVC-n] //
\endgl
\label{kwanhan-attr-mod1}
\xe

\ex
\small
\begingl
\gla 하늘\p{을} 확실히 \p{향한} 공//
\glb hanul-\p{ul} hwaksilhi \p{hyangha-n} kong //
\glc sky-ACC surely face-ADN ball//
\glft `a ball that is surely headed towards the sky' [PVC-p] //
\endgl
\label{hyanghata-attr-mod1}
\xe

\ex
\small
\begingl
\gla 명백히 게\p{에} \p{관한} 책 // 
\glb myengpaykhi key\p{-ey} \p{kwanha-n} chayk //
\glc clearly crab-OBL relate-ADN book  //
\glft `a book clearly about crabs' [PVC-n] //
\endgl
\label{kwanhan-attr-mod2}
\xe

In (\ref{hyanghata-attr-mod1}), insertion of an adverb inside what would have been a PVC/MWE still results in a grammatical sentence. This strongly points to an alternative analysis for this sentence, one in which \w{하늘}\p{을} \vpred{향한} \w{공 hanul}-\p{ul} \vpred{hyanghan} \w{kong} `ball towards the sky' forms a relative clause, as reflected in the translation. This would, conversely, mean that treatment of (\ref{hyanghata-attr-mod2}) could go either way, i.e., one that involves a PVC/MWE \textit{or} one with a relative clause.

\ex
\small
\begingl
\gla 확실히 하늘\p{을} \p{향한} 공//
\glb hwaksilhi hanul-\p{ul} \p{hyangha-n} kong //
\glc surely sky-ACC face-ADN ball//
\glft `a ball surely (headed) towards the sky' [PVC-p] //
\endgl
\label{hyanghata-attr-mod2}
\xe

\ex
\small
\begingl
\gla  친구\p{를} 용감하게 \p{구한} 강아지// 
\glb  chinku-\p{lul} yongkamhakey kuha-n kangaci //
\glc  friend-ACC courageously rescue-ADN puppy  //
\glft `a puppy who courageously rescued a friend' [verb+arg] //
\endgl
\label{kuhata-mod}
\xe

\paragraph{Connective verb modification.}
In a verb serialization construction, the final verb is the main predicate \citep{kim2010serial}. 
In this case, PVC verbs take the connective suffix to form PVCs, and cannot be modified (\ref{kwanhan-ser-mod}, \ref{hyanghata-ser-mod}).
Even non-predicative PVC verbs cannot take the predicative form due to their non-final verb position and remain lexicalized.
Similarly to (\ref{kwanhan-attr-mod2}, \ref{hyanghata-attr-mod2}), PVCs as a whole can be modified.
Non-PVC verbs can be freely modified (\ref{kuhata-ser-mod}).

\ex
\small\judge*
\begingl
\gla 게\p{에} 명백히 \p{관해} 서술한 책 // 
\glb key\p{-ey} myengpaykhi \p{kwanha-y} seswulhan chayk //
\glc crab-OBL clearly relate-CONN describe-ADN book  //
\glft `book written clearly about crabs' [PVC-n] //
\endgl
\label{kwanhan-ser-mod}
\xe

\ex
\small\judge*
\begingl
\gla 하늘\p{을} 확실히 \p{향해} 날아간 공//
\glb hanul-\p{ul} hwaksilhi \p{hyangha-y} nalakan kong //
\glc sky-ACC surely face-CONN fly-ADN ball//
\glft `ball flying while surely heading towards the sky' [PVC-p] //
\endgl
\label{hyanghata-ser-mod}
\xe

\ex
\small
\begingl
\gla  친구\p{를} 용감하게 {구해} 살린 강아지// 
\glb  chinku-\p{lul} yongkamhakey kuha-y salli-n kangaci //
\glc  friend-ACC courageously rescue-CONN save-ADN puppy  //
\glft `puppy who courageously rescued and saved the friend' [verb+arg] //
\endgl
\label{kuhata-ser-mod}
\xe

\end{document}